\documentclass{isprs}
\usepackage{subfigure}
\usepackage{setspace}
\usepackage{geometry} 
\usepackage{placeins}
\usepackage{multirow}
\usepackage{booktabs}
\usepackage{graphbox}
\usepackage{tabularx}
\usepackage{array}
\newcolumntype{L}[1]{>{\raggedright\let\newline\\\arraybackslash\hspace{0pt}}m{#1}}
\newcolumntype{C}[1]{>{\centering\let\newline\\\arraybackslash\hspace{0pt}}p{#1}}
\newcolumntype{R}[1]{>{\raggedleft\let\newline\\\arraybackslash\hspace{0pt}}m{#1}}

\usepackage{xcolor}
\definecolor{forest}{HTML}{009900}
\definecolor{shrubland}{HTML}{c6b044}
\definecolor{savanna}{HTML}{fbff13}
\definecolor{grassland}{HTML}{b6ff05}
\definecolor{wetlands}{HTML}{27ff87}
\definecolor{croplands}{HTML}{c24f44}
\definecolor{urban}{HTML}{a5a5a5}
\definecolor{snow}{HTML}{69fff8}
\definecolor{barren}{HTML}{f9ffa4}
\definecolor{water}{HTML}{1c0dff}

\usepackage{appendix}

\begin{document}

\title{Remote Sensing Image Classification with the SEN12MS Dataset}

\author{
 M.~Schmitt\textsuperscript{a,b}
 , Y.-L. Wu\textsuperscript{b}}

\address
{
	\textsuperscript{a}Department
of Geoinformatics, Munich University of Applied Sciences, Munich,
Germany - michael.schmitt@hm.edu\\
	\textsuperscript{b}Department of Aerospace and Geodesy, Technical University of Munich, Munich, Germany
}


\abstract
{
Image classification is one of the main drivers of the rapid developments in deep learning with convolutional neural networks for computer vision. So is the analogous task of scene classification in remote sensing. However, in contrast to the computer vision community that has long been using well-established, large-scale standard datasets to train and benchmark high-capacity models, the remote sensing community still largely relies on relatively small and often application-dependend datasets, thus lacking comparability. With this letter, we present a classification-oriented conversion of the SEN12MS dataset. Using that, we provide results for several baseline models based on two standard CNN architectures and different input data configurations. Our results support the benchmarking of remote sensing image classification and provide insights to the benefit of multi-spectral data and multi-sensor data fusion over conventional RGB imagery.
}


\maketitle

\section{Introduction}
One of the most crucial preconditions for the development of machine learning models for the interpretation of remote sensing data is the availability of annotated datasets. While well-established \textit{shallow} learning approaches were usually trained on small datasets, modern \textit{deep} learning requires large-scale data to reach the desired generalization performance. In computer vision, the great success of deep learning was largely driven by the desire to solve the image classification problem, i.e. assigning one or more labels to a given photograph. For this purpose, many researchers have relied on the ImageNet database \cite{Deng2009}, which contains millions of annotated images. In remote sensing, the same task is often called \emph{scene} classification, which similarly aims at assigning one or more labels to a remote sensing image, i.e., a scene. As \cite{Cheng2020} summarizes, there has also been a lot of progress in this field in recent years, with a growing number of dedicated datasets (cf. Tab.~\ref{tab:datasets}). As can be seen from this non-complete selection, most datasets built for remote sensing image classification deal with high-resolution aerial imagery, usually providing three or four spectral channels (RGB, or RGB plus near-infrared). Only EuroSat and BigEarthNet provide spaceborne multi-spectral imagery, with So2Sat LCZ42 being the only existing scene classification dataset covering the other large data modality -- synthetic aperture radar (SAR) data\footnote{After this paper was accepted for publication at ISPRS Congress 2021, it came to the authors' attention that BigEarthNet in the meantime was extended by BigEarthNet-S1, a collection of Sentinel-1 images corresponding to the Sentinel-2 images contained in the original dataset. Thus, now another multi-modal scene classification dataset exists.}. Combining all points, i.e. dataset size, availability of more than just a single sensor modality, and versatility, it becomes obvious that most existing datasets lack the power to train generic, region-agnostic models exploiting multi-sensor information.  

\begin{table*}
    \centering\footnotesize
     \caption{Non-exhaustive list of datasets for remote sensing image classification}
    \label{tab:datasets}
    \begin{tabular}{lllll}
    \toprule
    Dataset & Data Type & Number of Images & Image Size & Reference\\
    \cmidrule(r){1-1} \cmidrule(lr){2-2} \cmidrule(lr){3-3} \cmidrule(lr){4-4} \cmidrule(l){5-5}
    UC-Merced & Aerial RGB & $2{,}100$ & $256\times256$ & \cite{Yang2010}\\
    WHU-RS19 & Aerial RGB & $1{,}005$ & $600\times600$ & \cite{Xia2010}\\
    SIRI-WHU & Aerial RGB & $2{,}400$ & $200\times200$ & \cite{Zhao2016}\\
    NWPU-RESISC45 & Aerial RGB & $31{,}500$ & $256\times256$ & \cite{Cheng2017} \\
    AID & Aerial RGB & $10{,}000$ & $600\times600$ & \cite{Xia2017} \\
    PatternNet & Aerial RGB & $30{,}400$ & $256\times256$ & \cite{Zhou2018}\\
    EuroSat & Satellite multi-spectral& $27,000$ &$64\times 64$ & \cite{Helber2019}\\ 
    BigEarthNet& Satellite multi-spectral & $590{,}326$ & $120\times120$ & \cite{Sumbul2019}\\
    So2Sat LCZ42 & Satellite multi-sensor& $400{,}673$ & $32\times32$ & \cite{Zhu2020}\\
    \midrule
    \emph{SEN12MS} & \emph{Satellite multi-sensor} & \emph{180,662} & \emph{256~$\times$~256} & \emph{\cite{Schmitt2019}}\\
    \bottomrule
    \end{tabular}
\end{table*}

With this paper, we present the conversion of the SEN12MS dataset to the image classification purpose as well as a couple of baseline models including their evaluation. Since SEN12MS is -- in terms of spatial coverage and sensor modalities -- significantly larger than all other available datasets, and sampled in a more versatile manner, this will enhance the possibility to benchmark future model developments in a transparent way, and to pre-train remote sensing-specific models that can later be fine-tuned to individual problems and user needs. 

\section{SEN12MS For Image Classification}
In this section, the SEN12MS dataset in its new image classification variant is described. All resources, most notably labels or pre-trained baseline models, can be downloaded from 
\newline\texttt{https://github.com/schmitt-muc/SEN12MS} in an open access manner. The goal of both the repository and this paper is to support the establishment of standardized benchmarks for better comparability in the field.

\subsection{The original SEN12MS Dataset}
The SEN12MS dataset \cite{Schmitt2019} was published in 2019 and contains $180{,}662$ so-called patches, which are distributed across the world and all seasons. For each of those patches, the dataset provides, at a pixel sampling of 10m and a size of $256\times 256$ pixels,
\begin{itemize}
    \item a Sentinel-1 SAR image with two polarimetric channels (VV, VH)
    \item a Sentinel-2 optical image with 13 multi-spectral channels
    \item four different land cover maps following different classification schemes.
\end{itemize}

The SEN12MS dataset was designed having the following key features in mind:
\begin{itemize}
    \item Its main distinction from other deep learning-oriented datasets was (and is) its focus on multi-sensor data fusion. Instead of containing only optical imagery, SEN12MS provides both SAR and multi-spectral optical data to cover the most relevant modalities in remote sensing.
    \item Instead of being sampled over a single study area or a geographical region of limited extent (e.g. individual countries or continents) the data of SEN12MS is sampled from all inhabited continents. This makes the dataset unique with regard to the possibility to train generalizing models that are comparably agnostic with respect to target scenes.
    \item In contrast to its predecessor, the SEN1-2 dataset, all images contained in SEN12MS come as geotiffs, i.e. they include geolocalization information. On the one hand, this information can be used as an additional input feature (c.f. Uber's CoordConv solution \cite{Liu2018}). On the other hand, it can be used to pair the SEN12MS data with external geodata.
    \item By providing dense -- albeit coarse -- land cover labels for each patch, semantic segmentation for land cover classification was intended to be one of the main application areas of the dataset.
\end{itemize}
Since its publication, SEN12MS has been used in many studies on deep learning applied to multi-sensor remote sensing imagery. Examples include image-to-image translation \cite{Abady2020,Yuan2020} and land cover mapping with focuses put on weakly supervised learning \cite{Schmitt2020,Yu2020}, model generalization \cite{Hu2020}, and meta-learning \cite{Russwurm2020}. This shows the dataset's potential for both application-oriented as well as methodical research.

\subsection{Creation of Scene Labels from Dense Labels}
The original SEN12MS dataset contains four different schemes of MODIS-derived land cover labels. From those schemes, the IGBP scheme (see, e.g., \cite{SullaMenashe2019}) was chosen as background for the conversion into a classification dataset. This was done because the IGBP scheme features rather generic classes, including both natural and urban environments with a moderate level of semantic granularity. The other, LCCS-based, classification schemes, in contrast are less generic and over-focus on different topics of interest, e.g. land use or surface hydrology. As already proposed by \cite{Yokoya2020}, the $17$ original IGBP classes were converted to the simplified IGBP scheme (cf. Table~\ref{tab:IGBP_simple}) to ensure comparability to other land cover schemes such as FROM-GLC10 \cite{Gong2019}, and to mitigate the class imbalance of SEN12MS to some extent. 

\begin{table*}[!h]
\centering\footnotesize
\caption{The simplified IGBP land cover classification scheme.}
    \label{tab:IGBP_simple}
    \begin{tabular}{L{0.925cm} L{4.4cm} L{1.2cm} L{1.4cm} L{5.5cm} L{0.5cm}}
     \toprule
        IGBP Class Number & IGBP Class Name & Simplified Class Number & Simplified Class Name & Description & Color\\
        \cmidrule{1-1}  \cmidrule(lr){2-2} \cmidrule(lr){3-3} \cmidrule(lr){4-4} \cmidrule(lr){5-5} \cmidrule{6-6} 
        
        1  & Evergreen Needleleaf Forest & \multirow{5}{=}{1} & \multirow{5}{=}{Forest} & \multirow{5}{=}{Lands covered by woody vegetation at $>60\%$ and
height exceeding $2$~m} & \multirow{5}{=}{\textcolor{forest}{009900}}\\
        2  & Evergreen Broadleaf Forest & &    & \\ 
        3  & Deciduous Needleleaf Forest & & & \\
        4  & Deciduous Broadleaf Forest & & & \\
        5 & Mixed Forest & & &\\
        \addlinespace
        
        6 & Closed Shrublands & \multirow{2}{=}{2} & \multirow{2}{=}{Shrubland} & \multirow{2}{=}{Lands with shrub canopy cover $>10\%$ and $<2$~m tall} & \multirow{2}{=}{\textcolor{shrubland}{c6b044}}\\
        7 & Open Shrublands & & & \\
        \addlinespace
        
        8 & Woody Savannas & \multirow{2}{=}{3} & \multirow{2}{=}{Savanna} & \multirow{2}{=}{Lands with understory systems, and with forest
cover between $10\%$ and $60\%$ and $>2$~m tall} & \multirow{2}{=}{\textcolor{savanna}{fbff13}}\\
        9 & Savanna & & & \\\addlinespace\addlinespace
        
        10 & Grasslands & 4 & Grassland & Herbaceous lands with $<10\%$ trees/shrubs&  \textcolor{grassland}{b6ff05}\\\addlinespace\addlinespace
        
        11 & Permanent Wetlands & 5 & Wetlands & Lands with a permanent mixture of water and herbaceous or woody
vegetation& \textcolor{wetlands}{27ff87}\\\addlinespace\addlinespace
        
        12 & Croplands &  \multirow{2}{=}{6} & \multirow{2}{=}{Croplands} & \multirow{2}{=}{Lands covered with temporary crops followed by harvest and a bare soil period} &  \multirow{2}{=}{\textcolor{croplands}{c24f44}}\\
        14 & Cropland / Natural Vegetation Mosaics & & & \\\addlinespace\addlinespace
        
        13 & Urban and Built-up Lands & 7 & Urban/Built-up & Land covered by buildings and other man-made structures&  \textcolor{urban}{a5a5a5}\\\addlinespace\addlinespace
        
        15 & Permanent Snow and Ice & 8 & Snow/Ice & Lands under snow/ice cover throughout the year&  \textcolor{snow}{69fff8}\\\addlinespace\addlinespace
        
        16 & Barren & 9 & Barren & Lands with exposed soil, sand, rocks&  \textcolor{barren}{f9ffa4}\\\addlinespace\addlinespace
        
        17 & Water Bodies & 10 & Water & Oceans, seas, lakes, reservoirs, and rivers&\textcolor{water}{1c0dff}\\
    \bottomrule
    \end{tabular}
\end{table*}

\begin{figure*}[!h]
    \centering\footnotesize
   \begin{tabularx}{\linewidth}{XXXXXX}
    \includegraphics[width=0.9\linewidth]{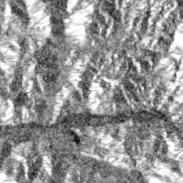}  & \includegraphics[width=0.9\linewidth]{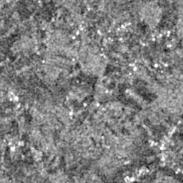} & \includegraphics[width=0.9\linewidth]{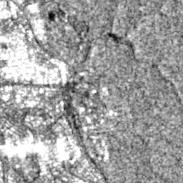}& \includegraphics[width=0.9\linewidth]{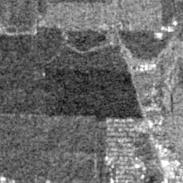}& \includegraphics[width=0.9\linewidth]{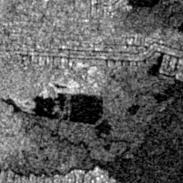} & \includegraphics[width=0.9\linewidth]{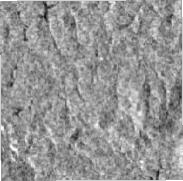} \\
   \includegraphics[width=0.9\linewidth]{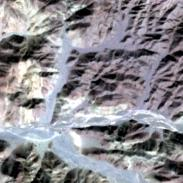}  &  \includegraphics[width=0.9\linewidth]{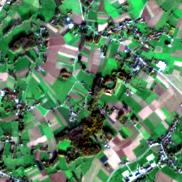}&   \includegraphics[width=0.9\linewidth]{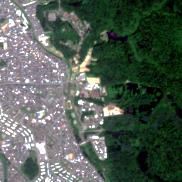}& \includegraphics[width=0.9\linewidth]{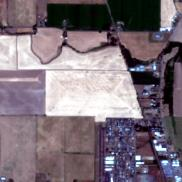}& \includegraphics[width=0.9\linewidth]{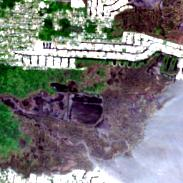} & \includegraphics[width=0.9\linewidth]{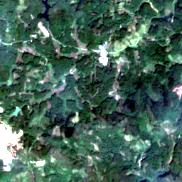}\\ 
      \textbf{Barren} & \textbf{Croplands} & \textbf{Urban/Built-Up}\newline Forest & \textbf{Croplands} & \textbf{Urban/Built-Up}\newline Savanna\newline Wetlands & \textbf{Savanna}
   \end{tabularx}
   \caption{Some randomly selected samples from the SEN12MS dataset, including simplified IGBP land cover annotations on scene level. First row: Sentinel-1 (VV backscatter). Second row: Sentinel-2 (RGB). The main class of the scene is set in bold, additional classes in the multi-label case are set in regular font.}\label{fig:examples}
\end{figure*}

For the generation of single-label scene annotations, simply the land cover class corresponding to the mode of the pixel-based land cover distribution in that scene was used. For the generation of multi-label scene annotations, the histogram of land cover appearances within a scene was converted to a probability distribution. Then, to remove visually underrepresented classes, only classes with a probability larger than $10\%$ were kept. An illustration of some example images including their single-label and multi-label scene annotations is provided in Fig.~\ref{fig:examples}.

\subsection{Dataset Statistics}
\begin{figure*}
    \centering
    \begin{tabular}{cc}
      \includegraphics[width=0.35\linewidth]{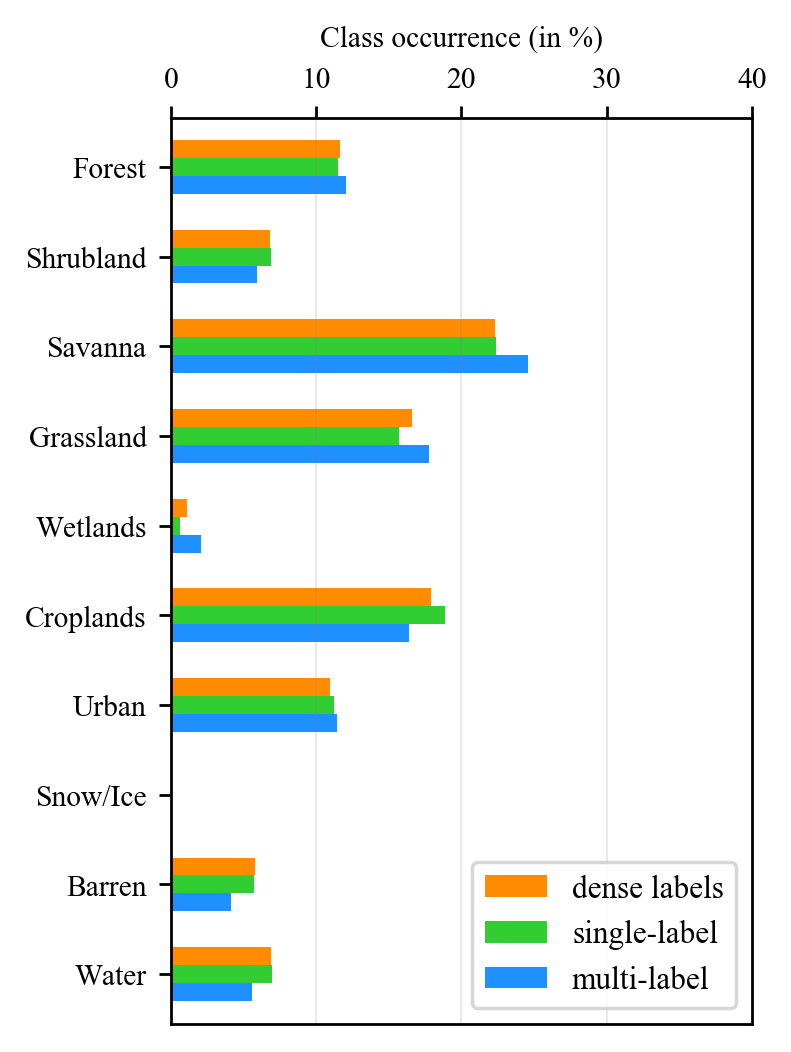}   & \includegraphics[width=0.35\linewidth]{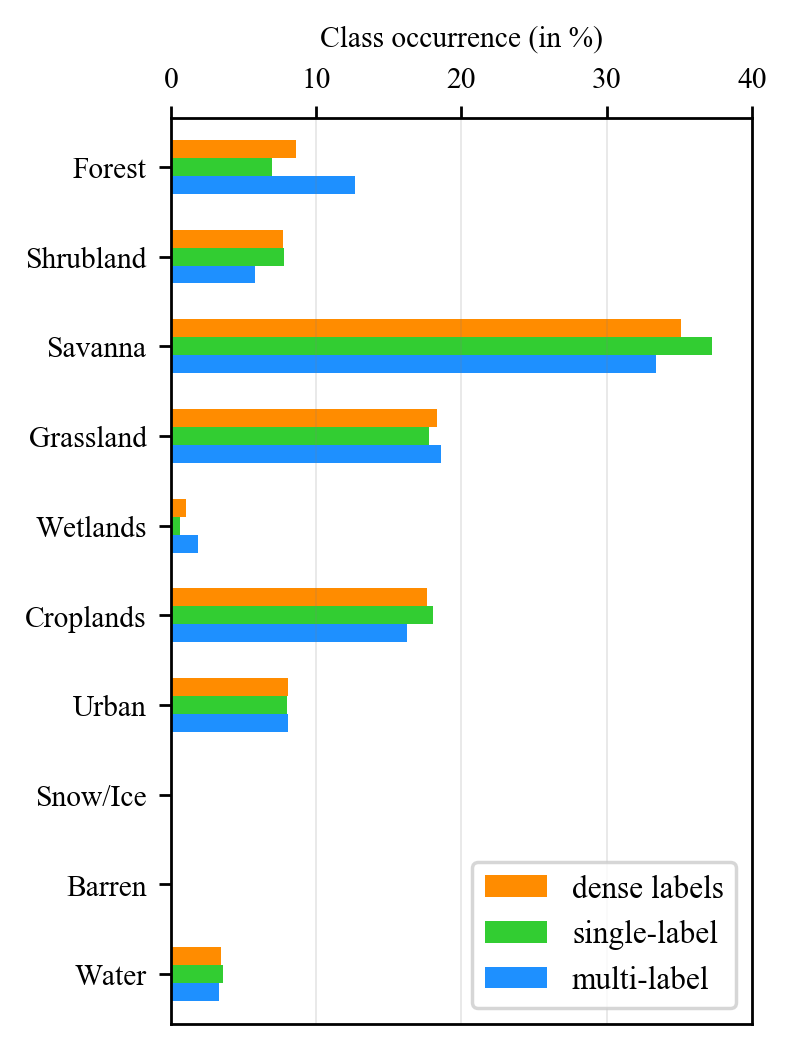}\\
        (a) & (b) 
    \end{tabular}
    \caption{Class occurrences in the different formats of the SEN12MS dataset. (a) Training split, (b) hold-out testing split. Note that the number of \emph{Barren} samples in the dense and the multi-label variant of the test set is actually not zero, but very small in comparison to the other classes.}
    \label{fig:class_occ}
\end{figure*}

The class distribution of the different label representations is summarized in Fig.~\ref{fig:class_occ}. It can be seen that the dataset is fairly imbalanced, with the classes \emph{Savanna}, \emph{Croplands}, and \emph{Grassland} being very frequent and classes such as \emph{Wetlands}, \emph{Barren}, and \emph{Water} being comparably underrepresented. The class \emph{Snow/Ice} can basically be considered as non-existing in SEN12MS. Due to the generally uniform spatial sampling of the SEN12MS data, this imbalanced class distribution is a representation of the natural imbalance of the real world, but should be considered when the data is used for the training and evaluation of machine learning models.

While \cite{SullaMenashe2019} provides a rough estimate of the accuracy of the original, global IGBP land cover map at $500$~m GSD -- namely $67\%$ --, there is no accuracy assessment of the upsampled IGBP maps provided as dense annotations in the original SEN12MS dataset. So, to provide a better intuition of the accuracy of the single-label and multi-label scene annotations presented with this paper, we have randomly selected $600$ patches for human annotation. This human annotation was conducted by visual inspection of the corresponding high-resolution aerial imagery in Google Earth. While this form of evaluation suffers from a certain subjectivity and the difficulty of distinguishing some land cover classes visually, it still serves the purpose to gain a better feeling for the label quality, if the numbers provided in Tab.~\ref{tab:label_val} are taken with a grain of salt. In this evaluation, a single human label was given to each patch, and Top-1 accuracy refers to the case in which the MODIS-derived single-label scene annotation matches this human label, whereas Top-3 accuracy refers to the case in which one of the top-3 multi-label scene annotations matches this human label. As can be seen, the average accuracy is somewhere around $80\%$, which is significantly better than the original accuracy of the global IGBP land cover map. This is, of course, caused by the simplification of the IGBP scheme, as well as the reduction to just few scene labels instead of coarse-resolution pixel-wise annotations. 
\begin{table}
    \centering
    \caption{Validation of the Scene Labels with Respect to Human Annotation}
    \label{tab:label_val}
    \begin{tabular}{lcrr}
    \toprule
         Class & Number & Top-1 Acc. & Top-3 Acc.\\
         \cmidrule{1-1}  \cmidrule(lr){2-2} \cmidrule(lr){3-3} \cmidrule{4-4}
         Forest & $66$ & $80.3\%$ & $97.0\%$\\
         Shrubland & $26$ & $46.2\%$ & $73.1\%$\\
         Savanna & $106$ & $88.7\%$ & $95.3\%$\\
         Grassland & $64$ & $68.8\%$ & $73.4\%$\\
         Wetlands & $3$ & $66.7\%$ & $66.7\%$\\
         Croplands & $162$ & $73.5\%$ & $84.6\%$\\
         Urban/Built-Up & $90$ & $80.0\%$ & $93.3\%$\\
         Snow/Ice & - & - & - \\
         Barren & $39$ & $92.3\%$ & $94.9\%$\\
         Water & $44$ & $97.7\%$ & $97.7\%$\\
         \midrule
         Average & $67$ & $77.1\%$ & $86.2\%$\\
         \bottomrule
    \end{tabular}
\end{table}

\section{Baseline Models for Single-Label and Multi-Label Scene Classification}
To provide the community with both baseline models and results, we have selected two well-established convolutional neural network (CNN) architectures for image classification: ResNet and DenseNet. The architectures and the necessary adaptations and settings are shortly described in the following. 
\subsection{ResNet}
The ResNet architecture \cite{He2016} was designed to mitigate the problem of vanishing gradients, which tended to appear for very deep CNNs before. This is realized by the introduction of so-called shortcut connections, i.e. instead of learning a direct mapping from input to output layer, the shortcut connection skips one or more layers by passing the original input through the network without modifications. Then, the network learns residual mappings with respect to this input. In this work, we used a ResNet50, i.e. a variant with a depth of 50 layers. 
\subsection{DenseNet}
The DenseNet architecture \cite{Huang2017} is based on the finding that CNNs can be deeper, more accurate and efficient to train if they contain shorter connections between layers close to the input and layers close to the output. Thus, DenseNets directly connect each layer to every other layer in order to ensure maximum information flow between layers in the network. Different to
ResNets, the features are not combined through summation; instead, they are concatenated. In this work, we employ a DenseNet121 with a depth of 121 layers.

\subsection{Training Details}
Both models were trained using binary cross entropy with logit loss for multi-label classification. To keep everything simple, optimization was performed with an Adam optimizer, a learning rate of 0.001, a decay rate of $10^{-5}$, and a batch size of 64. 

In order to evaluate the usefulness of different input data configurations, separate models were trained for the following cases:
\begin{itemize}
    \item S2\_RGB: Sentinel-2 RGB data only (as a computer vision-like baseline)
    \item S2\_MS: all 10 surface-related spectral channels of Sentinel-2
    \item S1+S2: Sentinel-1 dual-polarimetric data plus the 10 surface-related channels of Sentinel-2
\end{itemize}

Each model was trained from scratch on the official SEN12MS training split with early stopping based on a validation set randomly selected from the training set.

\section{Benchmark Results}
\begin{figure*}[h]
    \centering
    \begin{tabular}{lp{0.15\linewidth}p{0.2\linewidth}p{0.2\linewidth}}
     & \textbf{ResNet50} & \textbf{DenseNet121} & \textbf{\emph{Reference}}\\
     \cmidrule(lr){2-2} \cmidrule(lr){3-3} \cmidrule(l){4-4} 
      \includegraphics[align=c,width=0.15\linewidth]{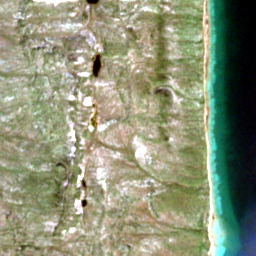} & Savanna\newline Wetlands\newline Water & Savanna\newline Wetlands\newline Water & \emph{Savanna}\newline\emph{Wetlands}\newline\emph{Water}  \\\addlinespace
      
        \includegraphics[align=c,width=0.15\linewidth]{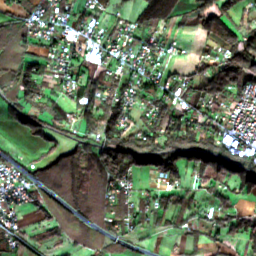} & Savanna\newline
Urban/Built-up & Savanna\newline
Croplands & \emph{Savanna}\\\addlinespace

        \includegraphics[align=c,width=0.15\linewidth]{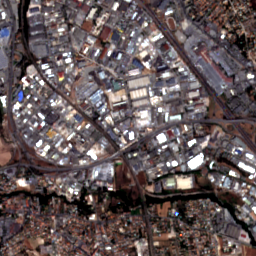} & Urban/Built-up & Urban/Built-up & \emph{Urban/Built-up}
    \end{tabular}
    \caption{Multi-label predictions for three example patches using the S1+S2 input data configuration. The second example is particularly noteworthy: While the reference denotes the patch as pure \emph{Savanna}, the models add finer information by recognizing urban or cropland structures, respectively.}
    \label{fig:pred}
\end{figure*}
Figure~\ref{fig:pred} illustrates multi-label prediction results for three example patches. 
Table~\ref{tab:multi_label_acc} contains a summary of accuracy metrics for the multi-label classification results. The F1 score is the harmonic mean of the precision and recall metrics, i.e.
\begin{equation}
F_1 = 2\frac{p\cdot r}{p+r},   
\end{equation}
where the precision $p$ is the fraction of correct predictions per all predictions of a certain class, and the recall $r$ is the fraction of correct predictions per all appearances of a class in the reference annotations.

\begin{table*}
    \centering
    \caption{F1 scores for the different models in the multi-label case.}
    \label{tab:multi_label_acc}
    \begin{tabular}{lcrrrrrr}
    \toprule
     & &\multicolumn{3}{c}{ResNet50}& \multicolumn{3}{c}{DenseNet121}\\
        Class & Number & S2\_RGB & S2\_MS & S1+S2 & S2\_RGB & S2\_MS & S1+S2\\
        \cmidrule(r){1-1} \cmidrule(lr){2-2} \cmidrule(lr){3-5} \cmidrule(l){6-8}
         Forest & $3529$ & $60.9$ & $69.3$ & $70.6$ & $65.4$ & $68.9$ & $70.6$\\
         Shrubland & $1611$ & $56.0$ & $50.2$ & $50.2$ & $58.3$ & $43.2$ & $45.0$\\
         Savanna & $9276$ & $79.3$ & $78.5$ & $79.9$ & $76.7$ & $80.2$ & $85.4$\\
         Grassland & $5171$ & $50.5$ & $58.4$ & $64.5$ & $50.9$ & $57.4$ & $60.0$\\
         Wetlands & $523$ & $11.0$  & $62.6$ & $55.8$ & $26.5$ & $53.7$ & $55.9$\\
         Croplands & $4511$ & $63.4$ & $66.9$ & $64.3$ & $66.3$ & $59.1$ & $63.2$\\
         Urban/Built-Up & $2235$ & $82.8$ & $82.6$ & $84.1$ & $84.1$ & $84.9$ & $83.1$\\
         Snow/Ice & - & - & - & -  &- &- &-\\
         Barren & $29$ & $1.3$ & $0.0$ & $11.5$  & $8.0$ & $0.6$ & $8.7$\\
         Water & $919$ & $70.4$ & $93.0$ & $93.7$ & $69.8$ & $93.8$ & $93.7$\\
         \midrule
         \textit{Average F1 Score} && $52.9$ & $62.4$ & $63.9$ & $56.2$ & $60.2$ & 62.8\\
         \textit{Overall F1 Score} && \textit{$66.5$} & \textit{$69.9$} & \textit{$71.4$} & \textit{$66.9$} & \textit{$68.9$} & \textit{$72.0$}\\
         \bottomrule
    \end{tabular}
\end{table*}

From the results, different insights can be drawn:
\begin{itemize}
    \item There is generally not a huge difference between the two baseline CNN architectures. This is particularly interesting, because both models are of significantly different depths. Thus, this provides a hint towards the hypothesis that the achievable predictive power is more limited by the training data than by the model capacity.
    \item Overall, the weakest performances are achieved for those models that take only optical RGB imagery as input. However, a measurable improvement is observed when multi-spectral data is used, with even more improvement provided by the data fusion-based model that exploits both optical and SAR data. This suggests that spectral diversity is of high importance in remote sensing-based land cover mapping and confirms once more that there is a significant difference between the analysis of conventional photographs and remote sensing imagery. 
    \item While SAR-optical data fusion is helpful on average, for some classes, e.g. \emph{Shrubland}, \emph{Wetlands}, and \emph{Croplands} it does not seem to be very helpful. This indicates that observation-level fusion based on a simple channel concatenation is not enough and more sophisticated fusion strategies, e.g. with sensor-dependent CNN streams, are needed. 
    \item Due to the reduction of dense labels to scene labels, the Barren class is underrepresented -- there are only 29 patches carrying a Barren label in the multi-label test dataset. This renders the results for the Barren class insecure. It is still interesting to note that for both CNN architectures, data fusion provides the best result for this class, while multi-spectral imagery yields the worst result -- even worse than RGB only.
    \item As already discussed in \cite{Schmitt2020}, the IGBP-based \emph{Savanna} label can be problematic: The MODIS-derived reference considers the scene of the second example in Fig.~\ref{fig:pred} as \emph{Savanna}, although it visually seems to rather be a mixture of built-up structures and croplands. Interestingly, both baseline models are able to identify those classes (i.e. \emph{Urban / Built-Up} for ResNet50 and \emph{Croplands} for DenseNet121) -- albeit not both at the same time, and without removing the \emph{Savanna} class. 
\end{itemize}
All in all, the achieved accuracies are of the same order of magnitude as the accuracies reported by \cite{Sumbul2019} for the BigEarthNet dataset. While SEN12MS and BigEarthNet are not directly comparable (as SEN12MS is more versatile and contains Sentinel-1 SAR imagery, but also uses a simpler class scheme), this indicates the usability of SEN12MS for the training and evaluation of scene classification models. Besides, the fact that achievable accuracies on both datasets are similar, this suggests a certain saturation for off-the-shelf image classification models for remote sensing scene classification. Investigating this further would be an interesting future research direction.

\section{Summary \& Conclusion}
With this paper, we have presented the SEN12MS dataset re-purposed for remote sensing image classification. To achieve this, the original land cover annotations, which are provided at a resolution of $500$~m and a pixel spacing of $10$~m, are converted to both single-label and multi-label annotations. Based on a randomized validation of $600$ patches by a human expert, an average accuracy of about $80\%$ was estimated. Using the dataset and two standard CNN image classification architectures, we have trained and evaluated several baseline models, which can serve as a baseline for future developments and provide insight to the benefit of using multi-sensor and multi-spectral data over plain RGB imagery.

\section*{Acknowledgments}
The authors would like to thank the \emph{Stifterverband} for supporting the ongoing work on the SEN12MS dataset and its derivatives with the Open Data Impact Award 2020. 

\bibliography{SEN12MS}

\end{document}